%% file: emnlp2023.tex
\pdfoutput=1
\documentclass[11pt]{article}

\usepackage{EMNLP2023}

\usepackage{times}
\usepackage{latexsym}

\usepackage[T1]{fontenc}
\usepackage[utf8]{inputenc}

\usepackage{microtype}

\usepackage{inconsolata}

\usepackage{booktabs}
\usepackage{graphicx}
\usepackage{multirow}
\usepackage{amsmath}
\usepackage{amssymb}
\usepackage{enumitem}
\usepackage{hyperref}
\usepackage[nameinlink]{cleveref}

\usepackage{verbatim}

\usepackage[T1]{fontenc}
\usepackage[utf8]{inputenc}

\usepackage{microtype}

\usepackage{svg}

\usepackage{todonotes}
\newcommand{\pj}[1]{\todo[color=yellow]{\small PJ: #1}} % Palak's comments
\newcommand{\lbs}[1]{\todo[color=green]{\small lbs: #1}} % Livio's comments
\newcommand{\tk}[1]{\todo[color=magenta]{\small tk: #1}} % Tom's comments

\renewcommand{\pj}[1]{{}} % Palak's
\renewcommand{\lbs}[1]{{}} % LBS
\renewcommand{\tk}[1]{{}} % Tom

\title{1-PAGER: One Pass Answer Generation and Evidence Retrieval}

\author{Palak Jain$^1$ \qquad Livio Baldini Soares $^2$ \qquad Tom Kwiatkowski$^2$ \\
$^1$ Google Research  \qquad  $^2$ Google Deepmind  \\
\texttt{\{palakj,liviobs,tomkwiat\}@google.com}}

\newcommand{\seal}{\textsc{SEAL}}

\newcommand{\locifull}{\textsc{1-Pager}}
\newcommand{\loci}{\textsc{1P}}

\begin{document}
\maketitle
\begin{abstract}
We present \locifull{} the first system that answers a question and retrieves evidence using a single Transformer-based model and decoding process.
\locifull{} incrementally partitions the retrieval corpus using \emph{constrained decoding} to select a document and answer string, and we show that this is competitive with comparable retrieve-and-read alternatives according to both retrieval and answer accuracy metrics. \locifull{} also outperforms the equivalent `closed-book' question answering model, by grounding predictions in an evidence corpus.
While \locifull{} is not yet on-par with more expensive systems that read many more documents before generating an answer, we argue that it provides an important step toward attributed generation by folding retrieval into the sequence-to-sequence paradigm that is currently dominant in NLP.
We also show that the \emph{search paths} used to partition the corpus are easy to read and understand, paving a way forward for interpretable neural retrieval.
\end{abstract}

\input{tex/introduction}

\input{tex/related_work_lbs}
\input{tex/approach_lbs}
\input{tex/inference_pj}

\input{tex/training}
\input{tex/experiments}

\input{tex/results}
\input{tex/discussion}
\input{tex/ablations}
\input{tex/conclusion}

\input{tex/limitations}

\input{tex/acknowledgement}
\input{tex/ethics}

\bibliography{anthology,custom}
\bibliographystyle{acl_natbib}

\clearpage
\appendix
\input{tex/appendix}

\end{document}

%% file: tex/introduction.tex
\section{Introduction}
In recent times, there has been a push to re-formulate a wide variety of tasks from NLP and other domains into the sequence-to-sequence paradigm, to make use of large pre-trained Transformer networks \cite{vaswani2017attention}.
However, despite evidence that large language models can often answer questions \cite{roberts2020much}, predict identifiers of documents that support those answers \cite{tay2022transformer}, or generate text that contains and explains an answer \cite{yu2022generate} the dominant paradigm in question answering is still the retrieve-and-read approach that pipelines separate retrieval and answer generation modules.
This approach has the benefit that it can provide direct and targeted paragraph-level attribution for the generated answers \cite{bohnet2022attributed}.
However, it also relies on a heterogenous mix of models that are hard to train in concert \cite{metzler2021rethinking}.
\input{fig/intro}

Motivated by the observation that language model decoders already perform search over possible sequences \cite{graves2012sequence}, and that evidence documents themselves are simply sequences of tokens, we present an alternative approach that relies on a single Transformer model.
In this approach, which we name \locifull{} (One \underline{P}ass \underline{A}nswer \underline{G}eneration and \underline{E}vidence \underline{R}etrieval) or simply \loci{}, the decoder iteratively partitions a corpus of evidence documents by generating a \emph{search path} consisting of a set of keywords that identify relevant documents and an answer string that is contained in at least one of these documents. With \loci{}, we aim to explore the spectrum between CBQA, where the answer is generated without reference to an evidence corpus, and pipelined approaches that feed retrieved documents into the transformer.

\Cref{fig:intro} illustrates an example in which the corpus is iteratively partitioned into documents that contain the string `Economy of India', then those that also contain the string `Agriculture', and finally those that also contain the answer string `23\%'.

\loci{} output sequences are guaranteed to match at least one document in the evidence corpus. This is enforced via a constrained decoder that has access to an FM-index representation of the evidence corpus contents \cite{892127} and we evaluate \loci{}'s ability to correctly answer open-domain questions while also retrieving passages that provide support for those answers \cite{bohnet2022attributed}.
Since \loci{} is the first model that can do both of these tasks, we compare to pipelined systems that first retrieve a single passage and then generate an answer based on this evidence passage.
\loci{} is competitive as a passage retriever, performing similarly to a widely used dense retriever \cite{karpukhin-etal-2020-dense} and outperforming the \seal{} system which independently generates keywords rather than a search path \cite{bevilacqua2022autoregressive}.
\loci{ } also outperforms an equivalent closed-book question answering (CBQA) model \cite{roberts2020much} according to answer accuracy.
Part of this improvement comes from the prediction of search paths themselves, reminiscent of chain-of-thought reasoning \cite{wei2022chain}, and part is from \loci{}'s constrained decoder, which forces the model to generate answers from passages that contain the keywords.

While \loci{} does not yet perform as well as the very best retrieval or open-domain question answering systems in terms of accuracy, the fact that it is competitive with pipelined systems that are trained with the same data and which use similar amounts of inference-time compute suggests a promising path ahead.
Unlike those systems, \loci{} can be trained end-to-end along with any other task that fits into the sequence-to-sequence paradigm. 
Additionally, \loci{} search paths are inherently interpretable, unlike embeddings used in dense retrieval.

%% file: fig/intro.tex
\begin{figure}[t]
\includegraphics[width=\linewidth]{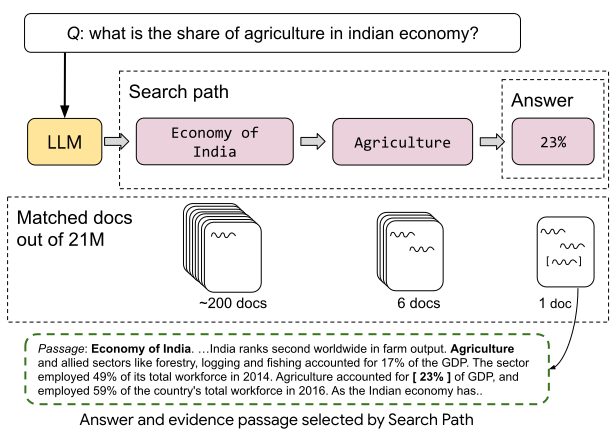}
% \includesvg[width=\linewidth]{fig/intro}
\caption{Example \loci{} output that iteratively partitions the corpus into sub-sets containing the generated n-grams. The last n-gram is taken as the answer.}
\label{fig:intro}
\end{figure}

%% file: tex/related_work_lbs.tex
\section{Related Work}
\input{fig/related_systems}

\paragraph{"Retrieve-and-read" Question Answering}

Question answering approaches in NLP are dominated by the "retrieve-and-read" paradigm  where a retriever first fetches hundreds of relevant documents from a corpus, followed by a language model that reranks and extracts the answer~\citep{harabagiu2003open,chen2017reading,zhu2021retrieving}. Sparse retrievers such as BM25~\citep{robertson2009probabilistic} build a high-dimensional lexical index over text corpus.
Dense retrievers~\citep{karpukhin-etal-2020-dense} use a dual encoder architecture to embed the query and document and perform an approximate nearest neighbor search. Various modifications to dense retrieval have been proposed over the years including hard negative training~\citep{xiong2020approximate}, late interaction~\citep{khattab2020colbert,santhanam2022colbertv2}, few-shot learning~\citep{izacard_few-shot_2022}, joint retriever and reader training~\citep{jiang2022retrieval}. 

A particular variant of interest is the \textbf{Iterative Retrieval} process where the query is reformulated incrementally~\citep{das2018multistep, lee2022generative} leading to an interactive search process~\citep{jiang2023active,adolphs2021boosting}. 
% qi-etal-2019-answering,min2019knowledge
This query augmentation scheme has similarities with our use of search paths. However, we use the paths to iteratively partition the corpus while prior works have used it for refining the query.

% \loci{} differs from retrieve-and-read approaches as it retrieves and extracts an answer span in a single pass of language model generation. To perform well, retrieve-and-read systems will typically retrieve 10s to 100s of passages that must be processed by a language model, making the process potentially several times more expensive.

To perform well, retrieve-and-read systems will typically retrieve 10s to 100s of passages that must be processed by a language model. In constrast, \loci{} retrieves and extracts an answer in a single pass of language model generation.

% \paragraph{Iterative Retrieval for Question Answering}

% One variant of "retrieve-and-read" model for question answering include \textit{iterative retrieval} models~\citep{das2018multistep,qi-etal-2019-answering,min2019knowledge, lee2022generative}, which are similar to \loci{} given that it also uses an iterative retrieval process. Beyond not requiring a separate reader model, another crucial difference between prior iterative retrieval approaches and \loci{} is that prior approaches use query rewriting/reformulation while we propose iterative corpus partitioning using the same query.

\paragraph{Closed Book Question Answering}

With data and parameter scale, LLMs in a closed-book setting (CBQA) have shown competitive performance~\citep{openai2023gpt4, anil2023palm,yu2023generate} to retrieval pipelines (ODQA), however without producing any attributed passages~\citep{rashkin2021measuring, bohnet2022attributed}. An extension of CBQA is post-hoc retrieval where a large language model LLM) is first used to generate an answer and then evidence for the question-answer pair is fetched by a retriever~\citep{gao2023rarr,bohnet2022attributed}. While post-hoc retrieval serves the same goal as \loci{}, it still uses a pipeline of LLM and retriever to do so.

% While post-hoc passage retrieval for question answering is conceptually similar to our iterative corpus partitioning retrieval, it differs in a few ways: (1) post-hoc retrieval requires separate models for question answering and retrieval, and (2) post-hoc retrieval is trained to retrieve solely with question and answer strings while our approach produces a  \textit{search path} that is semantically meaningful and often can guide the model to retrieve evidence for the answer.

% while we aim to perform the task in one pass with a single model.

\paragraph{Generative Retrieval}

% Recently, generative retrieval has been emerging as an alternative approach to the "retrieve-and-read" pipeline~\citep{metzler2021rethinking}. DSI~\citep{tay2022transformer} presented the first proof of LLM's capability to memorize document identifiers in the corpus. Other works have extended DSI by augmenting synthetic queries in training~\citep{wang2022neural}, building semantically richer document identifiers~\citep{zhou2022ultron} etc. However, memorization alone does not scale well to a corpus of million passages~\citep{pradeep2023does}.
% Semantically rich document representations are crucial to leverage generative powers of LLMs. Further to prevent hallucination, these generations need to be grounded in the

% Genre~\citep{decao2021autoregressive} employs constrained decoding to generate entities autoregressively from a pre-defined set. \seal{}~\citep{bevilacqua2022autoregressive} expands this constraint set to any keyword in the corpus 
% % with the use of FM-Index \citep{892127}, 
% facilitating a keyword-set to act as a document identifier. However, these systems still rely on a pipeline of re-ranker and reader to generate the final answer. 
% We take inspiration from and build on the above generative approaches to solve the task of passage-retrieval and question-answering in one go with a single language model.

Recently, generative retrieval has emerged as an alternative to the conventional "retrieve-and-read" pipeline~\citep{metzler2021rethinking}. Genre~\citep{decao2021autoregressive} performed generative entity linking by constraining model's decoding to a set of entities. DSI~\citep{tay2022transformer} showed one of the first proof of LLM's ability to memorize docids in the corpus. 
% Other works extend DSI by augmenting synthetic queries in training~\citep{wang2022neural}, ...
However, atomic ids or hierarchical clusters, as used in DSI, are opaque identifiers and capture limited information. Works such as \seal{}~\citep{bevilacqua2022autoregressive} and Ultron~\citep{zhou2022ultron}  use a semantically richer representation: keywords in the document. In particular, \seal{} constrains the generation to only keywords in the corpus using the FM-index \citep{892127}, a key data structure we borrow in this work.

\loci{} represents docids as keyword paths, which are arguably more interpretable, and learns a soft partition over the corpus instead of the hard partition imposed by DSI's clustering. 

Another crucial distinction is \loci{}'s ability to both retrieve and generate an answer while prior works rely on a external re-ranker/reader for the same. A high-level view of various question-answering systems is presented in \autoref{fig:related_systems}.

% However, memorization alone does not scale well to a corpus of million passages~\citep{pradeep2023does}.

% Document and query expansion using generative models~\citep{nogueira2019document,mao-etal-2021-generation,liu2022query}.

% Doc2Query \citep{nogueira2019document}: expands documents with synthetic queries.
% https://cs.uwaterloo.ca/~jimmylin/publications/Ma_etal_SIGIR2022.pdf
% \citep{liu2022query}

\paragraph{Attributed Question Answering}

Standard metrics for open-domain question answering, such as exact match or token-based F1, have received criticism for being imprecise and/or insufficient. Several efforts have proposed augmenting answers with textual evidence, via retrieval or citations~\citep{bohnet2022attributed,menick2022teaching,gao2023enabling}. While this work does not directly evaluate the quality of retrieved answer evidence, our proposed model inherently produces a passage to support the final answer, along with a search path of keywords, which could be used to provide users with answer evidence.

%% file: fig/related_systems.tex
\begin{figure*}[t]
\includegraphics[width=\textwidth]{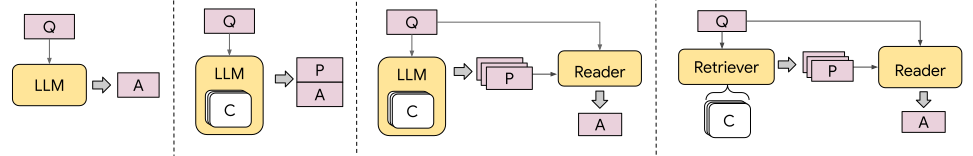}
% \includesvg[width=\textwidth]{fig/related_systems}
\caption{System illustration of different QA systems. From left to right: CBQA, \locifull{}, \seal{}, Retrieve-and-Read system. \textit{C} denotes the retrieval corpus, \textit{P} a retrieved passage, \textit{Q} the input question and \textit{A}, the generated answer. \loci{} is closest to CBQA (only single model used) but it also outputs a passage retrieved from \textit{C}.  
}
\label{fig:related_systems}
\end{figure*}

%% file: tex/approach_lbs.tex
\section{Iterative Corpus Partitioning and Answer Prediction}

% PJ commented: In this work, we aim to work on the attributed question answering task using generative language models. The attributed question answering task can be viewed as a combination of the open-domain question answering task and the passage retrieval task.
We focus on the problem of learning a mapping $f(q, D)\rightarrow (a, d_{a})$ from a question $q$ and corpus of documents $D$ to an answer and supporting document $(a, d_a)$. 
The predicted document $d_a$ is retrieved from $D$ and the answer $a$ is a sub-string of $d_a$.
The document $d_a$ should be relevant to the question and provide evidence for answer.
% predicting an answer and document pair $(a, d_{a})$, given a query $q$ and a set of documents , where $d_{a}$ is a passage that provides supporting evidence for the answer $a$, retrieved from a corpus of documents $D$.

The goal of this paper is to model the function $f$ using a single sequence-to-sequence model, rather than a pipeline which first retrieves $d_a$ and then feeds it into an answer generation module.
To achieve our goal, we recast retrieval as an \textit{iterative corpus partitioning} process illustrated in \Cref{fig:approach}.
% present a single system that predicts a 
% tackle the combination of open-domain question answering task and the passage retrieval task using generative language models.  
% In open-domain question answering, the task consists of producing an answer string $a$, given a query $q$. The passage retrieval task typically consists of, given a query $q$, returning a list of documents $d_{1}, d_{2}, ... , d_{n}$ from a retrieval corpus $D$. In the combination of these tasks we are interested in,
% % PJ commented: namely attributed question answering, 
% the system must produce a

\input{fig/approach}

% At a high-level, this involves generating a sequence of keywords (token n-grams) divided into two parts: (1) a sequence of keywords used for retrieval, and (2) an answer span contained within the retrieved passage(s). The sequence of keywords in the search path are used to iteratively partition the set of documents in the corpus, such that we select only documents that contain all generated keywords, resulting in a single (or few) documents. 

% PJ's alternate framing: At a high-level, this involves generating a sequence of keywords (token n-grams), divided into two parts (1) keywords used to iteratively partition the corpus and select a few documents (2) an answer span contained within the retrieved documents.

% Figure~\ref{fig:intro} illustrates the iterative retrieval process.
\lbs{Maybe go through the figure step by step here?}
\paragraph{Iterative corpus partitioning} adopts the LM decoder's autoregressive search process to partition $D$ by predicting n-gram \emph{keywords}.% that must be contained in the sub-selected set.

An n-gram of tokens $k$ is said to be contained in a document $d$, denoted by $k \prec d$, when $k$ is a sub-sequence of $d$. 
We define a \emph{keyword} corpus partitioning function
$$\mathcal{F}(D, k) = \{d | k \prec d; d \in D\}$$
that selects only those documents that contain $k$.
\locifull{} iteratively partitions the corpus $D$ by generating a sequence of n-grams that we refer to as a \textit{Search Path} $p_t = [k_{1}, k_{2}, \dots, k_{t}]$.
Each prefix of this search path defines a subset of $D$ via the \emph{search path} corpus partitioning function
$$\mathcal{P}(D, p_t) = D_{p_t} =\{ \cap_{i \in [1, t]} \mathcal{F}(D, k_i)\}$$
and each subsequent keyword $k_{t+1}$ narrows down $D_{p_t}$ into further sub-spaces such that $D_{p_{t+1}} \subseteq D_{p_t}$. 
%For example, in Figure~\ref{fig:approach} the search path iterativ consists of the three keywords 'Ten Commandments', 'twice in the Hebrew Bible', 'Book of Exodus' that identify a single passage out of 21m from Wikipedia.

% Finally, we produce a final n-gram ($k_{a}$) that corresponds to the answer string $a$ for the query $q$.
\paragraph{Answer prediction} is treated in exactly the same way as keyword selection and in \loci{} the last keyword from $p$ is taken as the answer.

% PJ commented: This formulation of the attributed question answering task has a few advantages. 
% This formulation of the task has a few advantages. 
\lbs{I think we should be explicit here about the pros of using this one-shot decode process.}
% Firstly, it enables us to generate the evidence and the answer together in a single shot. Secondly, retrieval as iterative corpus partitioning allows us to leverage language model's innate ability to generate  while the corpus guides the model to stay grounded at each step.

% The key idea is to generate search paths that incrementally narrow down the corpus to a small number of relevant documents. The reduced document space along with path-conditioning is expected to steer the model towards generating the optimal answer. 
% Moreover, the decoding mechanism allows us to automatically \textit{ground} and  \textit{attribute} the answer to the documents matched by the path. 

%% file: fig/approach.tex
\begin{figure*}[t]
\includegraphics[width=\textwidth]{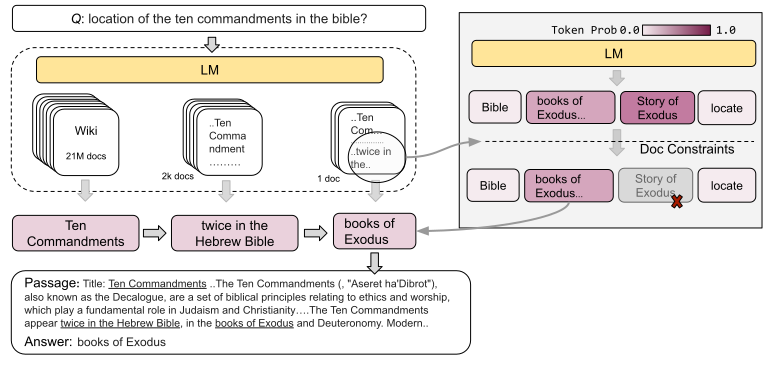}
% \includesvg[width=\textwidth]{fig/approach}
\caption{Illustration of the \loci{} decoding process. A keyword can only be generated from the documents matching previously generated keywords. Right panel shows a magnified view of applying constraints to a decoding step. Any keyword not present in the documents is masked out.}

\label{fig:approach}
\end{figure*}

%% file: tex/inference_pj.tex
\section{Constrained Decoding and FM-Index}
\label{sec:inference}
To avoid generating empty partitions, we constrain \locifull{} to only decode search paths that match at least one document.
We modify the decoder's beam-search strategy to only allow keyword continuations that are contained in the current partition.
% The decoder can only generate keyword continuations that are contained in the current partition.

%search paths that partition the corpus, we must constrain the language model to only decode tokens such that the resulting n-grams are always present in the corpus.
% \loci{} constrains the language model to only decode tokens where the corresponding ngrams are present in the corpus.
% For this, we rely on the FM-Index's ability to efficiently produce a set of allowed token continuations given a keyword prefix. 
Given a document subset $D_{p_i}$, which could be the full corpus $D$ at the start of decoding ($i=0)$, and a keyword prefix $k$, which could be empty, the set of all valid continuation tokens is defined as,
$$ \mathcal{C}(k, D_{p_i}) = \{ x | \hspace{5pt} k\!\mathbin\Vert\!x \prec d, d \in D_{p_i}\}$$ where $x$ is any vocabulary token and $\mathbin\Vert$ indicates concatenation of two token sequences. As a special case, when $k = \phi$ and $i = 0$, all tokens in $D$ are valid continuations.
\loci{} separates keywords in $p_T$ with a special separator token $\rightarrow$ and marks the end of the sequence with an {\sc eos} token. These two tokens are always valid continuations.

Consider \Cref{fig:approach}.
The three keywords correspond to the decoded token sequence [\emph{Ten, Commandments, $\rightarrow$, twice, in, the, Hebrew, Bible, $\rightarrow$, books, of, Exodus, {\sc eos}}].
At the start of decoding, any token in $D$ is allowed.
After decoding \emph{Ten}, only those tokens that follow \emph{Ten} as an n-gram in $D$ are allowed, along with the default separators.
After decoding [\emph{Ten}, \emph{Commandments}, $\rightarrow$] we are ready to start a new keyword, but only tokens from documents that contain the keyword \emph{Ten Commandments} are allowed.
Decoding continues in this manner until {\sc eos} is generated.
% At the start of decoding 
% At $X=\phi$, any token in $D_0$ can be generated. 
% For $X=$ "Ten", only tokens following "Ten" in $D_0$ are allowed. 
% For $X=$ "Ten", "Commandments" is a valid continuation since "Ten Commandments" occurs in $D_0$. 
% For $X=$ "Ten Commandments$\rightarrow$", any token in $D_\text{Ten\_Commandments}$ is permitted. 
% Next, "$\rightarrow$" is generated which is allowed all throughout decoding.  the second keyword is empty; next valid continuations can include any token in $D_\text{Economy of India}$. 
% Similarly at $X=$ "Ten Commandments$\rightarrow$", only tokens present in $D_\text{Ten\_Commandments}$ can be generated.
% % Similarly at $X=$ "Ten Commandments$\rightarrow$twice", only tokens following "twice" in $D_\text{Ten\_Commandments}$ can be generated.
% The decoding proceeds in this manner until {\sc eos} is emitted.  \pj{Do we need this example? Try trimming.}

To efficiently implement these constraints, we need a data-structure that can quickly determine both $\mathcal{C}(k, D_p)$, the continuation tokens given a document set and $\mathcal{P}(D_p, k)$, the subset of documents that contain a given path.

For this, we extend the usage of an FM-index~\cite{892127} as described in~\cite{bevilacqua2022autoregressive}. The FM-index is a compressed token-based index over a corpus $D_0$ with a few important properties for our usage: (1) it can efficiently list possible token continuations for a sequence prefix that occur in $D_0$ i.e., $\mathcal{C}(k, D_0)$, (2) it can list the set of documents in the corpus that match an n-gram i.e., $\mathcal{F}(D_0, k)$, and
(3) it supports search over arbitrary n-grams that occur \textit{within} documents. Note that the FM-index operations are optimized for $D_0$, the original corpus it is built over. We extend these to an arbitrary $D_p \subset D_0$ at additional cost described in \Cref{app:constrain compute}. 
% The modifications to decoding process to respect constraints is described in \Cref{sec:exp:inference}.

% The key idea is to generate search paths that incrementally narrow down the corpus to a small number of relevant documents. The reduced document space along with path-conditioning is expected to steer the model towards generating the optimal answer. Moreover, the decoding mechanism allows us to automatically \textit{ground} and  \textit{attribute} the answer to the documents matched by the path. 

%% file: tex/training.tex
\section{Training data generation}
\label{sec:training}

For training \loci{}, we produce a dataset with examples of queries and search paths as described above. 
% We derive such a dataset from existing open-domain question answering tasks that are grounded to gold passages, along with a retrieval corpus. 
At a high-level, we generate search paths by iteratively selecting n-grams from an answer passage, and simulating, using the FM-Index of the retrieval corpus, the partitioning of the corpus after selecting each keyword, until only a few documents remain. 
% We iterate keyword selection until the single gold passage (or a small set of documents which include the gold passage if the path is too long) and 
Finally, the answer span $a$ is appended to the search path. Each example produced can be serialized as sequence-to-sequence pair of inputs and targets as:

\begin{flushleft}
\texttt{inputs: Generate keywords for: <$q$>? \\
targets: \small{K\_SEP $k_0$ K\_SEP $k_1$ $...$ K\_SEP A\_SEP $a$ EOS}}
\end{flushleft}
% where  \texttt{\small{K\_SEP}} and \texttt{\small{A\_SEP}} are special tokens denoting keyword and answer separator respectively.

% Here we describe some approaches for heuristically sampling reasonable search paths that we use to train \loci{}. We provide evaluations and ablations of these in \cref{sec:discussion}.

\subsection{Keyword Selection}
A good keyword should have a) high relevance to the query and b) effectively narrow down the search space. To identify relevant keywords, we restrict to only the gold document $g$. 
All ngrams in $g$ of length up to five are extracted. Irrelevant keywords are filtered out such as those starting or ending with stop words. Similarly, keywords that are too rare in the corpus, e.g., "Philippines at Luzon" or too frequent, e.g., "part" are excluded based on a threshold on their count in corpus. The remaining keywords are scored with a combinations of heuristics, mainly  Rouge-1 similarity with the query \citep{lin2004rouge} along with minor award for keywords containing entities and penalty for keywords highly frequent in the corpus.
% \paragraph{Length of keywords} Ngrams beyond the length of 4-5 repeat rarely in the corpus. For instance, the keyword "Regional Rural Banks" occurs more than 200 times in Wikipedia corpus while "Regional Rural Banks were" occurs only once. Lengthy keywords could make the model generation go off-course fairly quickly. Thus, we prefer shorter ngrams of length upto 5 with majority of our keywords having a length of 2-3 words.
% A keyword such as "he is the" or "part of an" may provide very little information to resolve a query. To avoid such non-informative keywords, we filter out keywords that start/end with a stopword.

This scoring mechanism often misses out on keywords that are semantically relevant, but do not lexically overlap with the query. To boost the relevance of our keyword set, we re-score the top hundred keywords using a language model. A T5-XXL model is finetuned with the input as the query $q$ and target as either the title or a heuristically sampled keyword in a similar fashion to  \citet{bevilacqua2022autoregressive}. The heuristically sampled keywords are re-scored using this model to obtain a refined LM-scored set. Two other special types of keywords are awarded high scores: Title of the gold passage and the keyword containing the answer string $a$.

% To study the task's sensitivity to keywords,
% we also experiment with an noisy set of keywords $K_{noisy}$, sampled using the same heuristics as $K$, but without any pre-filtering i.e retaining keywords with trailing stopwords, rare-common keywords etc. 

% \subsection{Sampling Search Paths}
% Offline reinforcement learning formulations train on trajectories sampled from an expert model. \citep{chen2021decision, janner2021reinforcement}. For our initial model, we sample keywords based on its relevance and partitioning capabilities as described in \Cref{method:search path}.

\subsection{Search Paths}
The first keyword in a search path needs to effectively partition the corpus. We experiment with either the title or the highest scored keyword from the gold passage as the first keyword in the path.
The next keywords are sampled based on their score, given they do not overlap with any of the existing keywords in the path. We continue augmenting a path $p$ with keywords until at most ten passages in the corpus match i.e., $|D_p|$ < 10. The answer keyword is then appended to the path. Our train paths (including the answer) contain a median of three keywords and one matching document.
% and a special subset which we upsample, $T \rightarrow k_a$.

% \paragraph{Length of path} We observe that typically two salient keywords are sufficient to effectively partition the Wikipedia corpus and reduce the space to less than a dozen documents. For most of our experiments, we keep the path length between 2-3. This constrain gets automatically imposed since we truncate the path once the reward-to-go becomes 0.

% \paragraph{Path Redundancy} Consider a search path containing overlapping terms such as "Cannes Film Festival $\rightarrow$ Cannes Festival". Here the second term provides no new information. However, it may still increase the reward as the total number of documents containing both the keywords reduces. Such a reward path, may teach the model to simply emit repetitive terms. We filter out paths where more than half the tokens overlap between keywords.

%% file: tex/experiments.tex
\section{Experimental Setup}

\subsection{Datasets}
\label{sec:exp:datasets}
We use Open-NQ \citep{47761, lee-etal-2019-latent} as the question-answering dataset for training. For evaluation, besides Open-NQ, WebQuestions \citep{berant2013semantic} and CuratedTREC \citep{baudivs2015modeling} are used to measure out-of-domain performance.
The FM-Index corpus for constrained decoding is built over DPR Wikipedia corpus with 100-word splits ~\citep{karpukhin-etal-2020-dense}. The positive gold passages from DPR are used for sampling training paths.
This setup is chosen to mirror \seal{} and also permits fair comparison against DPR.

% We also experiment with unsupervised training data from PAQ \citep{lewis2021paq}, a dataset of 65M synthetically generated question-answer pairs from Wikipedia passages.

\subsection{Training}
\label{sec:exp:training}
% The training recipe involves two key ingredients: selecting relevant keywords for a query and building search paths. The search paths are then formulated as a seq2seq tasks using the format below:

% \begin{flushleft}
% \texttt{inputs: Generate keywords for: <$q$>? \\
% targets: \small{K\_SEP $k_0$ K\_SEP $k_1$ ..K\_SEP A\_SEP $k_a$ EOS}}
% \end{flushleft}

\loci{}'s training dataset contains 310k paths corresponding to 55k queries from Open-NQ. Majority of the training paths begin with the title, with a small fraction starting with other keywords~(12\%). All keywords, except the title, are scored using the LM-scoring technique described above. 

For our experiments, we use the T5X~\citep{roberts2022scaling} framework.
% including T5~\citep{raffel2020exploring} models, 
% which implements the Transformer-based ~\citep{vaswani2017attention} encoder-decoder architecture. Specifically,
A T5-XXL 1.1\footnote{\url{https://goo.gle/t5-checkpoints}}~\citep{raffel2020exploring} model is finetuned with a batch size of 256 and dropout of 0.1. No additional hyperparameter tuning is performed. We format search paths using the reserved tokens %in training examples we use reserved tokens from T5 vocab 
\texttt{\small{K\_SEP}} = "<extra\_id\_0>" and \texttt{\small{A\_SEP}} = "<extra\_id\_1>".

\subsection{Inference}
\label{sec:exp:inference}
% During decoding, beam search is modified to respect the corpus constraints. Given partial sequence $X_t=x_1x_2..x_t$ and corpus $D$, we compute the allowed continuations $\mathcal{C}(X_t, D)$ as described in \Cref{sec:inference}. All non-continuations are  masked out as illustrated in \autoref{fig:approach}.
% Hereafter, beam search proceeds in the standard manner.
% Thus, the model's original probability distribution for next token $\mathcal{M}(x_{t+1}|X_t)$ is modified to $\mathcal{M'}$ as follows:

% \begin{equation*}
% \small{
%     \mathcal{M'}(x_{t+1}|X_t)=
%     \begin{cases}
%     \mathcal{M}(x_{t+1}|X_t), & \text{if}\ x_{t+1} \in \mathcal{C}(X_t, D) \\
%      -\infty,&  \text{otherwise}\\
%     \end{cases}
%     }
% \end{equation*}

% \Cref{app:constrain compute} talks about the computational cost of the constraints and optimizations implemented.
Our best model employs beam decoding with a beam of 5. Even when the beam is greater than one, only the top-beam result is used for retrieval. We discuss the effect of beam size in depth in \Cref{sec:results}.
Given the top generated path $p$, $D_p$ corresponds to the retrieved documents. In case $|D_p| > 1$, a document is sampled arbitrarily for evaluation.
% The last keyword in $p$ is treated as the predicted answer. 

% The resulting paths match a single document for half the test set. The statistics of documents matched is depicted in \Cref{app:matched_docs}.
% To ensure the search path results in only a single document, we also experiment with forcing the model to decode the path until only a single  document matches and then decode the answer.

% \section{Experiments}\label{sec:
\subsection{Baselines}
\label{sec:exp:baselines}
We compare to a closed-book question answering (CBQA) system that generates answers, but does not ground these in an evidence corpus, as well as retrieve-and-read systems that combine a variety of retrievers with a Transformer-based answerer module.
Both the CBQA baseline and the answerer module are derived from the same T5-XXL 1.1 pretrained model as \loci{}.

% Since \loci{} is unique in outputting both a retrieved document and answer string, we c and predicting both answer and s one of the first systems that can both retrieve evidence retrieval and generate answer, we do not have a direct benchmark to compare. Instead, we evaluate two types of baselines: systems that generate answer without evidence (CBQA) and systems that only retrieve evidences. We opt for widely adopted dense retriever, DPR \citep{karpukhin-etal-2020-dense} and sparse retriever, BM25 \citep{robertson2009probabilistic} as baselines along with a generative retrieval approach, \seal{} \citep{bevilacqua2022autoregressive}. A Reader is attached in tandem to the retrievers for assessing question-answering performance.

\subsubsection{T5-CBQA}\label{sec:cbqa}
A T5-XXL 1.1 model is fine-tuned to predict answers from the DPR training set for 10,000 steps with a batch size of 128. 
% The hyperparameters batch size and drop-out were tuned on the validation set. 
Note that it is possible to achieve a higher closed-book performance on NQ using the full Open-NQ training split instead of the subset included in the DPR training set \citep{roberts2020much}. However, to enable meaningful comparison we restrict the CBQA baseline to the same training examples used to train \loci{}.

\subsubsection{Retrieve-and-Read}
\label{sec:baseline:retrieve-and-read}
% We compare to retrieve-then-read systems that combine a retriever with an answer generation module and report both retrieval and answer accuracy metrics.
The retrieve-and-read baselines first retrieve a \textit{single} passage from the evidence corpus, and then feed this passage and the question into the answer generation module\footnote{This differs from ODQA evaluations that do not include evidence retrieval as a first-class task, where many retrieved passages are fed into a reader that generates an answer without attribution to any single piece of text.}.
We report retrieval accuracy for the retrieved passage and answer accuracy for the generated answer.

\paragraph{T5-Reader} We tune a T5-XXL 1.1 model to generate answers from (question, evidence passage) pairs. This is the same base model used by \loci{} and we train on the (question, passage, answer) triples in the DPR training split to ensure fair comparison.
% based answerer, tuned on the DPR training set, with a retriever.

% combine a T5-XXL 1.1 model, tuned on the DPR training set, with a retrieval  and the retrievers 
% Since \loci{} combines both tasks in a single prediction, we report answer accuracy from the setting in which the answerer module is 
\paragraph{DPR-Retriever}
We compare against vanilla DPR finetuned on NQ without hard negatives \citep{karpukhin-etal-2020-dense} using the pre-computed index available on DPR's repository\footnote{\url{https://github.com/facebookresearch/DPR}}. 
We note that our ODQA setup differs from the one used by \citeauthor{karpukhin-etal-2020-dense} in that we choose the highest scoring retrieval as evidence for answer generation, instead of generating from the top-100 passages without attribution.

\paragraph{BM25-Retriever}
% For sparse retrieval baselines, we choose BM25 given it's widespread applications.
We use Pyserini toolkit~\citep{lin2021pyserini} with default configurations, retrieving the top-1 passage.

\paragraph{SEAL-Retriever}
\seal{} \citep{bevilacqua2022autoregressive} is a generative retrieval system that generates a set of keywords constrained on the corpus. In terms of technique, \loci{} borrows inspiration from \seal{}'s use of the FM-Index as well as keywords-as-identifiers.
However, the two setups have substantial differences that we highlight in \Cref{sec:discussion}.
We run \seal{} with its default configuration and a beam of 5 using the publicly released checkpoint based on Bart-large \citep{lewis-etal-2020-bart}. All outputs from the beam are used for retrieval.

\subsection{Evaluation}
\label{sec:evaluation}
We evaluate in-domain performance on the Open-NQ test split and out-of-domain performance on  WebQuestions (WQ) and CuratedTREC (TREC) following the setup from \citet{karpukhin-etal-2020-dense}. Passage retrieval performance is measured with Hits@1 using Pyserini evaluation scripts\footnote{\url{https://github.com/castorini/pyserini}}.
% Question answering accuracy is measured using Exact Match and F1 score. For \loci{}, the predicted answer is the last keyword of the generated path and, if the search path matches $>1$ passage, ties are broken randomly. For all other systems, the answer is generated from the top retrieved passage.
% following \citep{lee-etal-2019-latent, chen2017reading}.

\subsection{\loci{} configurations}
We experiment with three configurations:
a) \loci{}: Our primary setup that uses both training and constrained decoding procedures described above, producing a retrieved passage as well as an answer. b) \loci{}-Unconstrained: Only the training technique described in \Cref{sec:training} is adopted, with standard unconstrained decoding. Since generation is unconstrained, it is possible that no passage gets retrieved for a given path. c) \loci{}~+~Reader: Here, we take the top retrieved passage from \loci{} and input it to the Reader model (\Cref{sec:exp:baselines}) to extract the answer.

%% file: tex/results.tex
\section{Results}
\label{sec:results}
\input{tables/results}
\input{tables/results_ood}
We compare to the baselines described in \Cref{sec:exp:baselines} on Open-NQ using both retrieval and answer accuracy metrics in \autoref{tab:results}.
Answers are generated based on the top retrieved document in systems that separate retrieval from answer generation, to provide a clean comparison between systems that return (answer, evidence passage) pairs.
\autoref{tab:results_ood} reports the out-of-domain performance of various systems on WQ and TREC.

\loci{} outperforms CBQA in question answering and beats the retrieve-and-read systems, BM25 and \seal{}. 
On the passage retrieval task, it significantly improves over BM25 and \seal{}. 
% \loci{} is competitive with DPR in-domain and generalizes better out-of-domain.
% Overall \loci{} lags behind the QA pipeline that uses DPR but this appears to be more due to the reader rather than the retriever. We discuss this more in \Cref{sec:discussion}.
For in-domain setting, \loci{} is competitive with DPR on retrieval task, but lags behind the QA pipeline that uses DPR. However, this appears to be more due to the reader rather than the retriever as discussed in \Cref{sec:discussion}.
It is worth noting that \loci{} generalizes significantly better out-of-domain compared to other systems.

% In the rest of this section, we try to tease apart the role of training and decoding techniques.

\paragraph{Utility of Search Paths}
% The \loci{} Unconstrained setup adopts only the training technique described in \Cref{sec:training}, using the standard unconstrained decoding process. 
\loci{}-Unconstrained can be viewed as an extended version of CBQA that generates a search path before predicting the answer. Thus, improvement of \loci{}-Unconstrained over CBQA can be attributed to this path-conditioned answer generation process, analogous to chain-of-thought reasoning \citep{wei2022chain, lampinen2022can}. 
% Improvement of \loci{} Unconstrained over CBQA can, thus, be seen as an evidence of CoT on question-answering tasks.

\paragraph{Effect of Constrained Decoding}
\input{tables/planning_beam}
The purpose of constrained decoding is to ground the answer in an evidence retrieved from the corpus. As expected, the constrained setup enables \loci{} to achieve a higher Hits@1 than 1P-unconstrained.
% which has no visibility into the corpus contents. 
Surprisingly, when decoding with a beam of one, we observe a small drop in answer accuracy for \loci{} compared to \loci{}-Unconstrained (\autoref{tab:planning_beam}). Inspecting the losses, two dominant reasons surface. Firstly,  As DPR passages are chunked into 100-words \citep{karpukhin-etal-2020-dense}, some queries may become unanswerable given a single passage due to missing context. This is disadvantageous when the model has memorized the answer but there is no single passage to attribute it to.
% In the unconstrained setup, the model is free to generate any arbitrary keyword path and finally the correct answer. When constrained, it is forced to only pick an answer from a passage in the corpus. This is disadvantageous when the model has memorized the answer to a certain question but there's no single passage in the corpus to attribute it to. 

% Secondly, during constrained decoding, after generating the initial keywords, the model may end up in poor search spaces with no good candidates to pick from. 
% This is a typical problem with graph search which algorithms such as A* \citep{Hart1968} solve by back-tracking.
Secondly, during constrained decoding, after generating the initial keywords, the search space may soon become sparse with no good candidates to pick from. Could a larger room for planning its actions help the model here? Indeed, increasing the beam size to 5 improves performance by 3\% (\autoref{tab:planning_beam}), even when only the top-beam is used for retrieval. We refer to this as \textbf{Planning}, since the larger beam only enables the model to plan better and the remaining beam outputs are otherwise discarded. Note that unconstrained decoding does not gain from planning. 
% and not for the final evaluation.
In the final setup in \autoref{tab:results}, we use a beam of 5 for both \loci{} and \seal{}. 
Unlike \loci{}, \seal{} uses all the outputs from the larger beam for retrieval. 

%% file: tables/results.tex
\begin{table}[ht]
\begin{center}

\begin{tabular}{llcccc}
\toprule
\multirow{2}{*}{Retriever} & \multirow{2}{*}{Answerer} & Retrieval & \multicolumn{2}{c}{Answer} \\
 & & Hits @1 & EM & F1     \\ 
\midrule
-- & {\small T5 - CBQA} & -- & 26.8 & 34.0 \\
BM25 &  {\small T5 - Reader} & 23.6 & 17.9 & 24.0 \\
\seal{} & {\small T5 - Reader} & 37.9 & 29.4 & 35.8 \\
DPR  & {\small T5 - Reader} &  46.5 & 35.6 & 42.4 \\
\loci{} & {\small T5 - Reader} & 46.3 & 34.2 & 41.4 \\
\midrule
\multicolumn{2}{c}{\loci{} - Unconstrained} & 29.3 & 29.3 & 36.1 \\
\multicolumn{2}{c}{\loci{}} & 46.3 & 31.7 & 38.0 \\  
\bottomrule
\end{tabular}

\caption{Comparison of different Retriever and Answerer combinations on the NQ-Open test set. In retrieve-and-read setups, answers are generated from the top-1 retrieved passage. \loci{} combines passage retrieval and answer generation in a single prediction.}
\label{tab:results}

\end{center}
\end{table}

%% file: tables/results_ood.tex
\begin{table}[ht]
\begin{center}

\begin{tabular}{lcc|cc}
\toprule
\multirow{2}{*}{System} & \multicolumn{2}{c}{WebQuestions} & \multicolumn{2}{c}{TREC} \\
&  Hits @1 & EM &  Hits @1 & EM   \\ 
\midrule
BM25 + Rdr  & 19.7 & 14.2 &  35.2 & 29.1 \\
DPR \hspace{3pt} + Rdr  &  32.0 & 17.3 &  51.6 & 35.0 \\
\loci{} \hspace{11pt} + Rdr & 38.0 & 20.4 &  63.8 & 38.5 \\
\midrule
\multicolumn{1}{c}{\loci{}} & 38.0 & 20.5 &  63.8 & 36.4 \\  
\bottomrule
\end{tabular}

\caption{Comparison of different Retriever and Answerer combinations on Out-of-domain datasets. Both the Retriever and Answerer (Rdr) are trained on only Open-NQ. In retrieve-and-read setups, answers are generated from the top-1 retrieved passage.}
\label{tab:results_ood}

\end{center}
\end{table}

%% file: tables/planning_beam.tex
\begin{table}[ht]
\begin{center}

\begin{tabular}{l|c|c|c}
\toprule
\multirow{2}{*}{System} & Constrained   & \multicolumn{2}{c}{Beam}  \\
& Decoding & 1 &  5  \\
\midrule
CBQA & No & 26.7 &  26.8 \\
\loci{} Unconst. & No & 29.0 & 29.3 \\
\midrule
SEAL + Reader & Yes & 28.5 & 29.4 \\
\loci{} & Yes & 28.7 & 31.7 \\
\bottomrule
\end{tabular}

\caption{EM for various decoding setups with different beam sizes on Open-NQ. Only top-beam result is used for evaluation, except in \seal{} which uses all beam outputs. \loci{} constrained decoding benefits the most from a large beam whereas Unconstrained setups have only a slight effect.}
\label{tab:planning_beam}

\end{center}
\end{table}

%% file: tex/discussion.tex
\section{Discussion and Ablations}\label{sec:discussion}

\paragraph{Generating Answers}
% How does generating answers in a constrained mode fare to extracting answers from a passage? 
While \loci{} is capable of generating answers, \autoref{tab:results} highlights that it falls behind the \loci{}+Reader. The reason seems to be clear: the Reader has visibility into the full passage context while \loci{} is limited to the decoded search path and the constrained index which only ensures that generations are grounded in the corpus.
%The constrained setup only acts as a check ensuring the generations are grounded in the corpus.
% that can utilize the passage context.
% Besides \loci{} often emits incorrect answer entity type, eg: answering "who" questions with a date.
Since \loci{} does retrieve passages, it would be possible to pull in the 
%the passage ids are already retrieved during inference time, it is possible to provide the 
corresponding text as input for answer generation. 
%This effectively fuses in Reader within \loci{}. 
We leave this as future work.\pj{is this too far-fetched?}

\paragraph{Comparison to \seal{}}
While \loci{} takes inspiration from \seal{}, in practice, there are a few key differences between the two systems aside from \loci{}'s answer generation.

\seal{} generates a large set of keywords (\autoref{tab:seal_vs_loci}) using many separate decodes and heuristic guidance 
(\Cref{app:seal_keywords}). In contrast, \loci{} decodes a single sequence of about three keywords.

\input{tables/seal_vs_loci}

The \seal{} keywords are a set, decoded independently of each other and re-scored using sophisticated techniques to retrieve a large number of documents. For instance, the default configuration in \seal{} retrieves up to 500 documents. This makes \seal{} suitable to be employed in conjunction with a re-ranker.
In contrast, \loci{} search path's map directly to a single (or few) relevant documents (\Cref{app:matched_docs}).
% % We discuss the pros and cons of set vs path later in \Cref{sec:results}.

% Finally, like other retrievers, \seal{} does not generate an answer.
% % \loci{} generates the answer as part of its output and no further modeling is required. 
% % We use a Reader model to extract the final answer.

% \seal{} results are reported with its default configuration and a beam of 5 using the publicly released checkpoint based on Bart-large \citep{lewis-etal-2020-bart}. All outputs from the beam are used for retrieval.
We acknowledge the model-size variation between \seal{} and \loci{} in the reported experiments, however we preferred using the publicly available SEAL checkpoint. Given the discrepancies with larger beam-size, multiple decodes and use of Reader model, it is difficult to have an apples to apples comparison between the two systems.

% \paragraph{Reader}
% As mentioned earlier, all our baseline retrieval systems, namely DPR, BM25, \seal{}, only provide an evidence passage and no answer. A Reader model is applied on the top-retrieved passages to extract the final answer. A T5-XXL model finetuned on Open-NQ Reading comprehension task serves this purpose. 
% For completeness, we also report EM on answers extracted from the same Reader model on the \loci{} retrieved passages.
% A Reader module needs to used in tandem to extract the final answer. In our experiments, since we retrieve a single document, it suffices to use a standard Reading comprehension model for answer extraction.

\paragraph{Path vs Keyword set}
%\loci{} generates a path of 2-3 keywords that match a median of one document  while \seal{} an average of over 30 keywords, for a beam of 1. 
We qualitatively observe that keywords in a \loci{} path, owing to sequential generation, are distinct and add new information as compared to the \seal{} output set where overlapping keywords are common (\Cref{app:seal_keywords}). 
Thus, paths are advantageous for precisely narrowing down to a single relevant document while keyword sets are effective for retrieving a large number of documents that can later be reranked. 
This is corroborated by the fact that \loci{} is better at Hits@1 while \seal{} is better at Hits@5 (\Cref{app:hits5}).
% When retrieving the top-passage (Hits@1), paths  outperform a much larger keyword set. On the other hand, \seal{} achieves a higher Hits@5 accuracy than \loci{} as noted in \Cref{app:hits5}, since the latter rarely retrieves 5 or more documents (\Cref{app:matched_docs}).

% Note that for a beam of 5, \seal{} uses all keywords generated from the beam to retrieve documents while  \loci{} uses the beam to accommodate planning and only the top-beam result is used for retrieval.

% Keyword set can be repetitive while keywords path are likely to be unique owing to conditional generation. 

\input{tables/examples}

\paragraph{Qualitative Analysis}
\autoref{tab:examples} illustrates patterns of Search Paths generated by \loci{}. We note some of the common path patterns here:

1) First keywords are entities in the query, followed by query predicates that iteratively narrow down towards an answer. This is the most common type of path observed and can be attributed to the dominant presence of title in the training data.

2) Rewrites of the original query or related predicates such as "seasons consists of",  "appeared on~...". Such paths are more prevalent where there is no canonical entity in the query or no entity can be determined with high confidence.
    
3) Answer is directly generated followed by supporting keywords that guide towards an attributed passage. This happens in a small fraction of cases, likely where the pretrained model has memorized an answer with high confidence.

Overall, we find the generated search paths to be fairly meaningful and interpretable.

%% file: tables/seal_vs_loci.tex
\begin{table}[ht]
\begin{center}

\begin{tabular}{c|c|c}
\toprule
 & \seal{}  & \loci{} \\
\midrule
% Median keywords  & \multirow{2}{*}{32} & \multirow{2}{*}{3} \\
% generated & & \\
Median keywords & 32 & 3 \\
Median docs retrieved & ~500 & 1 \\
Generates answer & $\times$ & $\checkmark$ \\
\bottomrule
\end{tabular}

\caption{Key differences between \seal{} and \loci{} measured over Open-NQ test split with a beam of 1.}
\label{tab:seal_vs_loci}

\end{center}
\end{table}

%% file: tables/examples.tex
\begin{table*}[ht]
\begin{center}
\small
\begin{tabular}{p{0.7\textwidth}|cp{0.2\textwidth}}
\toprule
 Query (Q) and Generated Search Path (SP) & Comment \\
\midrule
\multicolumn{2}{l}{\textbf{\textit{Correctly attributed passages and answers}}} \\
\midrule
~Q: how many episodes of greys anatomy season 14 & Query entity resolved first,\\
SP: Grey's Anatomy (season 14) >> season consists of 24 episodes >> 24  &  followed by query predicates \\
\midrule
~Q: when did they start adding zinc to pennies	&  Query entity resolved  \\
SP: Penny (United States coin) >> zinc >> Lincoln cent >> 1943 & iteratively \\
\midrule
~Q: who was executed for being an american spy during the revolutionary war & \multirow{2}{*}{Answer generated first} \\
SP: Nathan Hale >> Army during the American Revolutionary >> Nathan Hale &   \\
\midrule
~Q: who was the grandfather on the cosby show & \multirow{2}{*}{Query rewrites} \\
SP: appeared on "The Cosby >> Earle Hyman	&  \\
\midrule
\midrule
\multicolumn{2}{l}{\textbf{\textit{Incorrect Passage or Answer}}} \\
\midrule
~Q: who decides the number of judges in the high court \hspace{40pt} A: President of India	&  Path correctly resolved, \\
SP: judge is appointed >> High Court >> Chief Justice of India	& Failed on answer\\
\midrule
~Q: when did the isle of wight become an island \hspace{50pt} A: During the last Ice Age & Query entity resolved, \\
Isle of Wight >> 1890 >> 1890  & Failed on supporting keywords\\
\midrule
~Q: love yourself by justin bieber is about who \hspace{110pt} A: Rihana & Failed to resolve \\
SP: Love Yourself: Her >> music video >> Her  & query entity\\
\midrule

\bottomrule
\end{tabular}

\caption{Example \loci{} Search Paths (SP) on Open-NQ test set. The last keyword in SP is the predicted answer. Gold answers are indicated by A.}
\label{tab:examples}

\end{center}
\end{table*}

% Correct examples
% Q: location of the ten commandments in the bible &	\\
% SP: Ten Commandments >> twice in the Hebrew Bible >> books of Exodus and Deuteronomy	& \\
% \midrule
% Q: what category was hurricane charley when it hit florida & Correct entities \\
% SP: Hurricane Charley >> Category >> 2004 Atlantic hurricane season >> Category 4 &  generated \\
% Q: what is the share of agriculture in indian economy & \\ 
% SP: Economy of India >> Agriculture and allied sectors >> 23\% & \\
% \midrule
% Q: who is the book of acts written to	& \\
% SP: Acts of the Apostles >> Jewish audiences >> Jewish audiences, &\\
% Q: form from material that has accumulated on the earths surface & \\
% SP: formed by weathering >> Sedimentary rock >> sedimentary rock & \\
% \midrule
% Q: where can the mona lisa be found today	& \\

% Incorrect Examples
% Q: who had a baby at 100 in the bible \hspace{5pt} (A: Sarah,Abraham) & Failed to resolve \\
% SP: Methuselah >> According to the Book >> Methuselah & Key entity \\
% Q: what is the maximum data rate for the 802.11a standard select one \hspace{20pt} A: 54 Mbit/s & Path correctly resolved,\\
% SP: 802.11a >> data rate >> 2.4 GHz &  Failed on answer \\
% \midrule

%% file: tex/ablations.tex
% \section{Ablations}
% \label{sec:ablations}

\input{tables/keyword_paths}
\paragraph{Sampling Search Paths for Training}
\autoref{tab:keyword_paths} highlights that high quality keywords are crucial to performance. The LM re-scored set of keywords result in significant accuracy gain over heuristically sampled keywords.
Paths with first keyword as Title boost performance further. Mixing in a small fraction of paths starting with non-title keywords encourages the model to generate predicates where no entity can be determined, giving us the best results.
% as also observed in \citep{decao2021autoregressive, bevilacqua2022autoregressive}.

\paragraph{Sensitivity to tokenization}
We find that constrained decoding is highly sensitive to rare tokenization or punctuation formatting in the corpus. Consider the query "who sang i ran all the way home" with the gold document title "Sorry (I Ran All the Way Home)". In the unconstrained setup, the model's top prediction  starts with "I Ran All the Way Home". However, "(I" is tokenized differently from "I" and searching over the FM-Index returns no match. As a result, constrained decoding drops the predicted keyword altogether, resorting to lower ranked keywords in the beam.
We partially fix the issue by modifying the answer in a fraction of the training data to include surrounding punctuation tokens based on how they appear in the FM-index.
% ensuring the model is sensitive to such tokens during training. A fraction of answer keywords in the training dataset are updated to include the surrounding punctuation tokens based on how they appear in the FM-Index.
For instance, the keyword "I Ran ..." would update to "(I Ran ...".
This simple change leads to a jump in answer accuracy from $26.4\%$ to $28.7\%$. However, much more work is needed to make \loci{} robust to variations in tokenization.
% studying the effect of tokenization and mitigation strategies. 
% \autoref{tab:tokenization} notes the non-trivial jump in accuracy from this seemingly minor fix. Another potential solution may be inference time Token Healing \footnote{https://github.com/microsoft/guidance\#token-healing-notebook}. 
% change affects about 10\% of the examples in the training set,
%  The improvement becomes less pronounced for larger planning beam where there is room for back-tracking.
% indicating that lower ranked outputs may already include keyword variations with punctuation or articles. 

% Updating all keywords in the training set to respect corpus tokenization could further alleviate the issue. 

See \Cref{app:datasize} for analysis of training data size and \Cref{app:calibrating} for masking logits vs logprobs.
% and \Cref{app:force_title} for force-decoding title.

%% file: tables/keyword_paths.tex
\begin{table}[ht]
\begin{center}

\begin{tabular}{l|c|c}
\toprule
Search Path & Hits@1 & EM \\
\midrule
% $k_h \rightarrow k_h$ & 34.5 & 22.6 \\
% $k_{lm} \rightarrow k_{lm}$ & 40.0 & 27.2 \\
% $t \rightarrow k_{lm}$ & 41.9 & 28.0 \\
% $t \rightarrow k_{lm} +$ & \multirow{2}{*}{42.9} & \multirow{2}{*}{28.7} \\
%  $k_{lm} \rightarrow k_{lm}~(7+1)$  & & \\
Heuristic & 34.5 & 22.6 \\
LM-{\small scored} & 40.0 & 27.2 \\
Title >> LM-{\small scored} & 41.9 & 28.0 \\
% LM-{\small scored} + Title >> LM-{\small scored} & 42.9 & 28.7 \\
Title >> LM-{\small scored} + & \multirow{2}{*}{42.9} & \multirow{2}{*}{28.7} \\
LM-{\small scored}~(7+1)  & & \\
\bottomrule
\end{tabular}

\caption{Comparison of Training Search Paths on Open-NQ. Here LM-{\small scored} denotes re-scoring by LM on a heuristic set. All results are with a beam of one. ">>" indicates keyword separator and "+"  mixture of path types in the give ratio.}
\label{tab:keyword_paths}

\end{center}
\end{table}

%% file: tex/conclusion.tex
\section*{Conclusion}
We introduce \locifull{}, the first system to perform question answering and passage retrieval in one pass with a single language model, using a constrained decoder to iteratively partition the retrieval corpus and then generate an answer. 
We show competitive or improved performance over a variety of comparable baselines and carefully analyze the results, ablating both training strategies and decoding style. We also provide a qualitative analysis of predictions to illustrate the system's capabilities. Challenges with constrained decoding are surfaced including poor search spaces and sensitivity to tokenization and mitigation strategies are presented.

We hope that 1P adds value in demonstrating how a single transformer model can be harnessed to do both retrieval and  answering and pave the path for further progress in the generative retrieval domain.

%% file: tex/limitations.tex
\section*{Limitations}
\loci{} is geared towards identifying a concise, small set of documents and generating answer in a single go. While this makes the architecture simpler, it also adds certain weaknesses. \loci{} is not effective for retrieving a large number of documents and falls behind pipelined systems that combine retrieval with re-ranking. Even for a single passage, it lags behind state-of-the-art dense-retrieval techniques. \loci{}'s method of answer generation is also not competitive with the use of a reader, due to lack of passage context. 

Our training strategy relies heavily on titles or entities and it's generalization on corpora without rich structure or on queries without central entities, remains to be studied.

Constrained decoding also comes with its own challenges. Constrained beam outputs often lack diversity, so that even with a larger beam one may still end up in poor search spaces. Computing document-level constraints across the corpus is expensive as it may require scanning a large number of rows in the index. Further, communication between FM-Index and Transformer model slows down inference.

% Further, FM-index needs to be placed on a CPU since current TPU implementations do not support certain index operations. The communication between TPU and CPU proves to be a major bottleneck leading to very large inference time.

%% file: tex/acknowledgement.tex
\section*{Acknowledgement}

We thank Don Metzler,  Nicholas FitzGerald, Partha Talukdar, Srini Narayanan, as well as our anonymous reviewers, for their thoughful comments and valuable feedback

%% file: tex/ethics.tex
\section*{Ethical Considerations}
While Large Language Models can solve a wide range of tasks effectively, they also suffer from biases across axis such as gender, race, region  \citep{chan2023gpt}. LLMs are also prone to generating toxic content, especially when probed about it. Although, our task grounds the model's generations on a corpus, some of the biases in pre-trained LLMs, may seep in \locifull{}.

Building the FM-index and constrained decoding is a compute-intensive affair. We have experimented over a single dataset, Natural Questions, involving only knowledge-seeking queries, and single model family, T5. It is possible that some of our findings may not hold over other datasets or model families. Finally, our experiments are limited to English corpus and queries. The proposed approaches are resource-intensive and may not be accessible or valid for several low-resourced languages.

%% file: tex/appendix.tex
\section{Appendix}

\subsection{Constrain Computation}
\label{app:constrain compute}
% FM-Index is a compressed suffix-array based index optimized for locating sub-strings in a corpus and finding their continuations.
\loci{} relies on two key operations for constrain computation:
\begin{enumerate}[label=\alph*)]
    \item $\mathcal{F}(D, k)$ : Documents that contain keyword $k$
    \item $\mathcal{C}(k, D)$ : Next tokens for keyword $k$ in arbitrary document set $D$
\end{enumerate}

$\mathcal{F}(D, k)$ is preprocessed and cached to allow for quick computation. $\mathcal{C}(k, D)$ is trickier to compute. When D represents the full corpus, FM-index can fetch the next tokens in  $O(|V|log(|V|))$, where $V$ is the token vocabulary and independent of $|D|$. However, arbitrary $D$ requires a traversal over all documents and can be very expensive. In practise, the LLM training guides it to generate effective keywords such that  $|D|$ is small.
% Thus, for fast computations it is essential that either the keyword itself is highly effective i.e. $|D_k|$ is small or the preceding path is precise i.e. $|D|$ is small.
% Keywords such as "the" or "this" match a large set of documents and therefore, can lead to a very large computation time. During training time, we guide the model to generate effective keywords.

% Similarly, the cost of computing the corpus partition $\mathcal{F}(D', k)$ is proportional to  $min(|D_k|, |D'|)$.

% $O(|D_k|*log(N))$ where $N$ is the number of documents in $D_0$.

% Thus, for fast computations it is essential that either the keyword itself is highly effective i.e. $|D_k|$ is small or the preceding path is precise i.e. $|D'|$ is small.
% Keywords such as "the" or "this" match a large set of documents and therefore, can lead to a very large computation time. We must be careful during training to avoid  keywords with a high $|D_k|$.

We also apply certain other optimizations to reduce the compute cost: \begin{itemize}
    \item Constrains are computed lazily over a decoding pass.
    \item Several computations are cached, eg: keyword to document id mapping
    \item To cap the cost of constraints at each decoding step, we allow for unconstrained generation in rare scenarios, when the estimated cost is too high. If the generated path is absent in the corpus (<1\% examples), these can be filtered out later.
\end{itemize}

%  and documents cached during decoding. In rare scenarios, we also allow unconstrained generation for a time-step when computing $C(k, D)$ is estimated to be too expensive. Such outputs can be post-filtered if they do not occur in the corpus. 

Despite these optimizations, inference continues to be expensive and we perhaps need a special data structure for next token look-up.

\subsection{Training data size}
\label{app:datasize}
\input{tables/datasize}
In \autoref{tab:datasize}, we observe the effect of dataset size on performance. Increasing the numbers of paths sampled per query improves performance, perhaps due to higher diversity in training. However, this method of dataset expansion is limited by the number of relevant paths we could extract for a query. 

We also experiment with increasing the query set manifold by mixing in unsupervised datasets. A total of 9M QA pairs are sampled from PAQ \citep{lewis2021paq}, a synthetic QA dataset, and search paths extracted with heuristic scoring described in \Cref{sec:training}. The original \loci{} training dataset is mixed in 1:1 ratio. This further boosts performance, but not proportionally to the amount of data added, indicating diminishing returns from silver datasets. 

% Augmenting the training data with 30 times the queries from an unsupervised source gives us only marginal gains indicating diminishing returns in using silver data that may be lower in quality than the Open-NQ gold data.

% First we keep the query set fixed, using
% the question-answers from the Open-NQ  train splits \citep{karpukhin-etal-2020-dense} consisting of 58,800 samples. 
% and increase the number of paths manifold. 
% We attribute the resulting gain to better generalization capabilities. 

\subsection{SEAL keywords}
\label{app:seal_keywords}
\input{tables/seal_vs_loci_examples}
\seal{} generates a set of document substrings constrained on the corpus, that are combined to form document identifiers.
Besides using a LM to generate keywords, \seal{} utilizes several other mechanisms for extracting keywords. This includes partial beam sequences, heuristically adding query n-grams, sampling the top-k tokens from the logprobs of the first decoding step, force decoding title etc. The keywords are re-scored using the LM as well as FM-index count and all keyword combinations are retrieved. 
\autoref{tab:seal_vs_loci_ex} illustrates keywords generated by both the systems. Note that \seal{} keywords can be repetitive and therefore require a large number of keywords to narrow down to meaningful documents.
This also makes \seal{} suitable for retrieving a much larger set of documents that can be re-ranked later.  The maximum number of retrieved documents for \seal{} are capped by a hyperparameter with default value of 500. In contrast, \loci{} is geared towards retrieving only the top-document.

\subsection{Hits@5}
\label{app:hits5}
\input{tables/hits5}
\seal{} does significantly better than \loci{} for Hits@5 (\autoref{tab:hits5}). We attribute this to the large set of keywords generated by \seal{} as explained in the \Cref{app:seal_keywords}.

\subsection{Normalizing sequence likelihood over constrained space}
\label{app:calibrating}
During constrained decoding a sequence $X$, we need to choose the next token from $\mathcal{C}(X, D)$  and not the entire vocabulary space $V$. Should the sequence likelihood be re-normalized over this constrained space? We find that re-normalizing the probabilities results in inflated likelihoods, making it hard for the model to back-track.

Consider the query,  "where did the butchers in the slaughterhouse cases live" to which our model predicts an irrelevant search path [\emph{Slaughterhouse Five}, \emph{but}, \emph{\sc EoS}]. What's going on under the hood? The first keyword is incorrect lending the model into a poor search space. With the second keyword, the model is possibly looking to generate "butcher" but there's no such keyword in the constrained set. Ideally, the model should backtrack at this point to other candidates in the beam. However, since the set of continuations is small, re-normalizing  inflates the probablities of all tokens in $\mathcal{C}$ including \emph{EoS}, even though the true likelihood of such a sequence is very low.
Indeed, using the language model's scores directly without any re-normalization cures this issue yielding [\emph{Slaughterhouse cases}, \emph{Butcher}, \emph{\sc EoS}]. and this is the strategy we opt for in all our experiments.

% \subsection{Force decoding Title keyword}
% \label{app:force_title}
% In \autoref{tab:keyword_paths}, we note that title is a critical keyword in generating relevant paths. Our training data biases the model to predict a title-like first keyword, but the model has no external knowledge of the titles in Wikipedia. We experiment with constraining the first keyword to belong to the set of Wikipedia titles, allowing the next keywords to be produced following regular constraints. This actually causes a drop in accuracy indicating that title may not always be a good first keyword and imposing such structures on the output space may hurt the model.

\subsection{Number of matching documents}
\label{app:matched_docs}
\input{fig/matched_doc}
\loci{} generated paths effectively narrow down the corpus, generally matching only a few documents in the corpus as illustrated in \autoref{fig:matched_doc}. Note that a small fraction of paths match 0 documents due to pruning optimizations applied during inference time detailed in \Cref{app:constrain compute}.

%% file: tables/datasize.tex
\begin{table}[ht]
\begin{center}

\begin{tabular}{l|c|c|c|c}
\toprule
Dataset & Queries & Paths & Hits@1 & EM \\
\midrule
Open-NQ & 55k & 55k & 41.9 & 28.1  \\
Open-NQ & 55k & 310k & 42.9 & 28.7 \\
Open-NQ & 55k  & 310k & \multirow{2}{*}{43.6} & \multirow{2}{*}{29.5} \\
+ PAQ & + 9M & + 9M & & \\
\bottomrule
\end{tabular}

\caption{Comparison of different dataset sizes for queries and paths}
\label{tab:datasize}

\end{center}
\end{table}

%% file: tables/seal_vs_loci_examples.tex
\begin{table*}[ht]
\begin{center}
\small
\bgroup
\def\arraystretch{1.2}
\begin{tabular}{p{0.05\textwidth}|p{0.65\textwidth}lp{0.15\textwidth}} \\
\toprule
System & Question or Search Path & Answer  \\
\midrule
& who has the most catches in nfl history & Jerry Rice \\ 
\loci{} & 2,000-yard club >> Barry Sanders & Barry Sanders\\
\seal{} & </s> Michael Irvin @@, yards per catch, caught his, touchdown, record & T.J. Houshmandzadeh\\ 
\midrule
& when was harry potter and the philosophers stone published & 1997 \\
\loci{} & Harry Potter and the Philosopher's Stone >> first published in the United >> 1997 & 1997\\
\seal{} &  </s> Harry Potter and the Philosopher's Stone @@, "Harry Potter, Potter and the Philosopher's Stone is, Potter and the Philosopher's Stone Harry, novel & 1999\\ 
\midrule
& what is the meaning of the harp in ireland &  the arms of Ireland \\ 
\loci{} & Harp >> national symbol of Ireland >> national symbol of Ireland & national symbol of Ireland\\
\seal{} & </s> Harp @@, Irish harp,, harp is, harp was, harp &  aristocracy\\ 
\midrule
& who was the president of pakistan during 1971 war &  Yahya Khan \\ 
\loci{} & Indo-Pakistani War of 1971 >> Prime Minister of Pakistan >> Zulfikar Ali Bhutto& Zulfikar Ali Bhutto\\
\seal{} & </s> Indo-Pakistani War of 1971 @@, East Pakistan, Pakistani, Pakistan Army, Pakistan's &  Muhammad Yaqub Khan\\ 
\midrule
& when do you declare honors in contract bridge &  any time after the auction \\ 
\loci{} & Contract bridge >> declaring >> end of the hand & end of the hand \\
\seal{} & </s> Contract bridge @@, declarer, bidding, honors, hands &  bidding\\ 
\bottomrule
\end{tabular}
\egroup

\caption{Comparison of keywords generated by \seal{} and \loci{} for randomly sampled exampled from Open-NQ test set. For \loci{}, we show  the full search path separated by ">>" with the last keyword as the answer. For \seal{}, we illustrate the top-5 keywords along with the answer from Reader model. "</s>" and "@@" are special tokens used by \seal{} for identifying start of passage and title marker respectively. The Answer next to the question is the gold answer while others are predictions from corresponding systems.}
\label{tab:seal_vs_loci_ex}

\end{center}
\end{table*}

%% file: tables/hits5.tex
\begin{table}[ht]
\begin{center}

\begin{tabular}{l|c|c}
\toprule
System & Beam  & Hits@5 \\
\midrule
\seal{} & 1  & 59.7 \\
\seal{} & 5  & 62.8 \\
\loci{}  & 1  & 46.5 \\
\loci{}  & 5 & 50.8 \\
\bottomrule
\end{tabular}

\caption{Hits@5 on Open-NQ test. \seal{} achieves a much higher score than \loci{} owning to the larger number of documents matched and re-scored. Note that only top-beam result is used for \loci{} while \seal{} uses all beam outputs.}
\label{tab:hits5}

\end{center}
\end{table}

%% file: fig/matched_doc.tex
\begin{figure}[t]
\includegraphics[width=\linewidth]{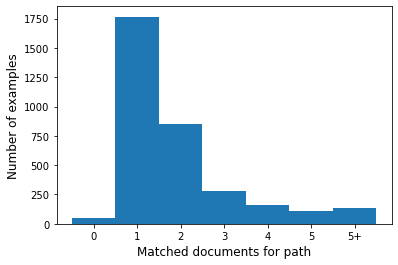}
\caption{Number of matching documents in the corpus for \loci{} generated path in the test set. About half the examples match only a single path.}
\label{fig:matched_doc}
\end{figure}